\documentclass{article}
\pdfoutput=1
\usepackage{spconf,amsmath,graphicx}
\usepackage{amssymb}
\usepackage{multirow}
\usepackage{setspace}

\title{Exploring Pre-training with Alignments for RNN Transducer based End-to-End Speech Recognition}

\name{Hu Hu$^{1,2}$\sthanks{The work was done as an intern at Microsoft.}, Rui Zhao$^{1}$, Jinyu Li$^{1}$, Liang Lu$^{1}$, Yifan Gong$^{1}$}
\address{$^1$Microsoft Speech and Language Group, Redmond, WA, USA \\
$^2$Georgia Institute of Technology, Atlanta, GA, USA}

\begin{document}
\ninept
\maketitle

\begin{abstract}
Recently, the recurrent neural network transducer (RNN-T) architecture has become an emerging trend in end-to-end automatic speech recognition research due to its advantages of being capable for online streaming speech recognition. However, RNN-T training is made difficult by the huge memory requirements, and complicated neural structure. A common solution to ease the RNN-T training is to employ connectionist temporal classification (CTC) model along with RNN language model (RNNLM) to initialize the RNN-T parameters. In this work, we conversely leverage external alignments to seed the RNN-T model. Two different pre-training solutions are explored, referred to as encoder pre-training, and whole-network pre-training respectively. Evaluated on Microsoft 65,000 hours anonymized  production data with personally identifiable information removed, our proposed methods can obtain significant improvement. In particular, the encoder pre-training solution achieved a 10\% and a 8\% relative word error rate reduction when compared with random initialization and the widely used CTC+RNNLM initialization strategy, respectively. Our solutions also significantly reduce the RNN-T model latency from the baseline.
\end{abstract}


\vspace{0.1cm}
\section{Introduction}
\label{sec:intro}
In recent years, we have witnessed significant progress in automatic speech recognition (ASR) mainly due to the use of deep learning algorithms \cite{speech-dnn1, speech-dnn2}. Deep model based ASR systems mainly focus on the hybrid framework and consist of many components, including acoustic model (AM), pronunciation model, language model (LM). Those components are trained separately using different objective functions, and extra expert linguistic knowledge may be needed. Recently, an emerging trend in the ASR community is to rectify this disjoint training issue by replacing  hybrid systems with end-to-end (E2E) systems \cite{sak2015learning, miao2015eesen, las, prabhavalkar2017comparison, chiu2018state, battenberg2017exploring, sak2017recurrent, hadiantowards, sainath2018improving, li2018advancing}. The three major E2E approaches are built on the Connectionist Temporal Classification (CTC) \cite{ctc-graves, e2e-graves, ctc-ms}, Attention-based Encoder-Decoder (AED) \cite{Cho-RNNEncDecSMT, Bahdanau-RNNEncDecAlignTranslate, bahdanau2016end, Chorowski-AttentionASR, las}, and recurrent neural network transducer (RNN-T) \cite{rnnt-graves, rnnt-google, rnnt-ms}. Different from training conventional hybrid models, token alignment information between input acoustic frames and output token sequence is not required when training the E2E models.

CTC maps the input speech frames to target label sequence by marginalizing all the possible alignments. A dynamic programming based forward-backward algorithm is usually used to train the model. An advantage of the CTC approach is that it does frame-level decoding as the conventional hybrid model, and hence can be applied for online speech recognition. However, one disadvantage is the conditional independence assumption given the input acoustic frames. AED, on the other hand, does not have such an assumption, and is presumably more powerful than CTC for the speech recognition task. However, one drawback of the AED model is that the entire input sequence is required to start the decoding process due to the global attention mechanism, which makes it challenging for real-time streaming ASR, despite some recent attempts along this direction~\cite{chiu2017monotonic, moritz2019triggered}. 

RNN-T is an extension to CTC, which consists of three components: an encoder, a prediction network, and a joint network which integrates the outputs of encoder and prediction networks together to predict the target labels. RNN-T overcomes the conditional independence assumption of CTC with the prediction network; moreover, it allows streaming ASR because it still preforms frame-level monotonic decoding. Hence, there has been a significant research effort in promoting this approach in the ASR community \cite{rnnt-ms, rnnt-google, rnnt-google-multilin, rnnt-baidu, graves2013speech}, and RNN-T has recently been successfully deployed in embedding devices~\cite{rnnt-device}. 

However, compared to CTC or AED, RNN-T is much more difficult to train due to the model structure, and the synchronous decoding constraint. Besides, its training is very memory demanding due to the 3-dimensional output tensor \cite{rnnt-google, rnnt-ms}. In \cite{rnnt-ms}, an approach is proposed to reduce the memory cost, and it enables large mini-batch training. To tackle the training difficulty, initializing the encoder and prediction networks of an RNN-T with a CTC model and an RNNLM respectively is proven to be beneficial~\cite{rnnt-google, graves2013speech}. In this paper, we explore other model initialization approaches to overcome the training difficulty of RNN-T models. Specifically, we propose to utilize external token alignment information to pre-train RNN-T. Two types of pre-training methods are investigated, which are referred to as encoder pre-training and whole-network pre-training respectively. Encoder pre-training refers to initializing the encoder in the RNN-T only, while the other components are trained from the random initialization. The whole-network pre-training, as its name suggests, pre-trains the whole network by an auxiliary objective function instead of the RNN-T loss. The proposed methods are evaluated on 3400 hours voice assistant data and 65,000 hours production data. The experimental results show that the accuracy of RNN-T model can be significantly improved with our proposed pre-training methods, with up-to 28\% relative word error rate (WER) reduction.

The rest of this paper is organized as follows: Section~\ref{sec:rnnt} briefly introduces the basic RNN-T model, including the model training and decoding. The proposed two types of pre-training methods are described in Section~\ref{sec:init-sep} and Section~\ref{sec:init-com}, respectively. Next, Section~\ref{sec:exp} shows the experimental results and analysis. Section~\ref{sec:con} gives the conclusions.

\section{RNN Transducer Model}
\label{sec:rnnt}
\begin{figure}[t]
  \centering
  \includegraphics[width=0.6\linewidth]{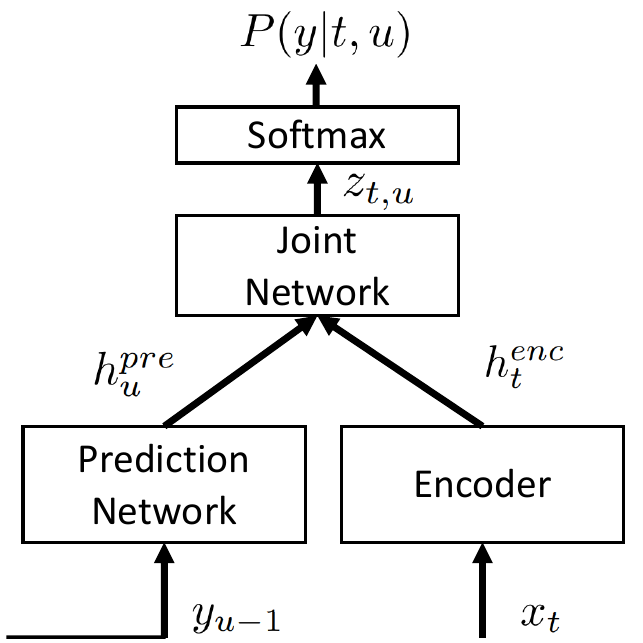}
  \vspace{0.0cm}
  \caption{RNN-Transducer model structure.}
  \label{fig:rnnt}
\end{figure}

The RNN-T model was proposed in \cite{rnnt-graves} as an extension to the CTC model. A typical RNN-T model has three components, as shown in the Figure~\ref{fig:rnnt}, namely encoder, prediction network and joint network. Compared with CTC, RNN-T does not have the conditional independence assumption because of the prediction network, which emits an output tokens conditioned on the previous prediction results. 

To be more precise, the encoder in an RNN-T model is an RNN that maps each acoustic frame $x_t$ to a high-level feature representation $h^{enc}_t$, where $t$ is the time index:
\begin{equation}
h_t^{enc} = f^{enc} (x_t).
\end{equation}
The prediction network, which is also based on RNNs, converts previous non-blank output token $y_{u-1}$ to a high-level representation $h_u^{pre}$, where $u$ is the label index of each output token.
\begin{equation}
h_u^{pre} = f^{pre} (y_{u-1}).
\end{equation}

Given the hidden representations of both acoustic features and labels from the encoder and prediction network, the joint network integrates the information using a feed-forward network as:
\begin{align}
z_{t,u}  &= f^{joint} (h_t^{enc}, h_u^{pre}).
\end{align}

The posterior probability $P(y|t, u)$ can be obtained by taking the Softmax operation on the output of the joint network. Then a forward-backward algorithm \cite{rnnt-graves} is performed on the three-dimension output to compute the total probability $P(\textbf{y}|\textbf{x})$ of the output sequence $\textbf{y}$, conditioned on the input sequence $\textbf{x}$. The negative log-loss of the target sequence is used as the objective function to train the model, 
\begin{align}
L_{RNN-T} = -log\ P(\bf{y}|\bf{x}).
\end{align}

The decoding of RNN-T is operated in a frame-by-frame fashion. Starting from the first frame fed to encoder, if the current output is not $blank$, then the prediction network is updated with that output token. Otherwise, if the output is $blank$, then the encoder is updated with the next frame. The decoding terminates when the last frame of input sequence is consumed. In this way, real-time streaming is satisfied. Greedy search and beam search can be used in the decoding stage, which stores different numbers of intermediate states.

\section{Encoder Pre-training}
\label{sec:init-sep}
In an RNN-T model, encoder and prediction network usually have different model structures, which make it difficult to train them well at the same time. Directly training RNN-T from random initialization may get a biased model toward one of the model components, i.e., dominated by the acoustic or the language input. Most groups adopt a initialization strategy that initializes the encoder with a CTC model and the prediction network with a RNNLM \cite{graves2013speech, rnnt-google, rnnt-xielei}. However, the output sequence of CTC is a series of spikes, separated by the $blank$ \cite{ctc-graves}. Thus after CTC based pre-training, most encoder output $h^{enc}_t$ leads to generate $blank$, which results in  wrong inference for the RNN-T model. 

\begin{figure}[ht]
  \centering
  \includegraphics[width=0.8\linewidth]{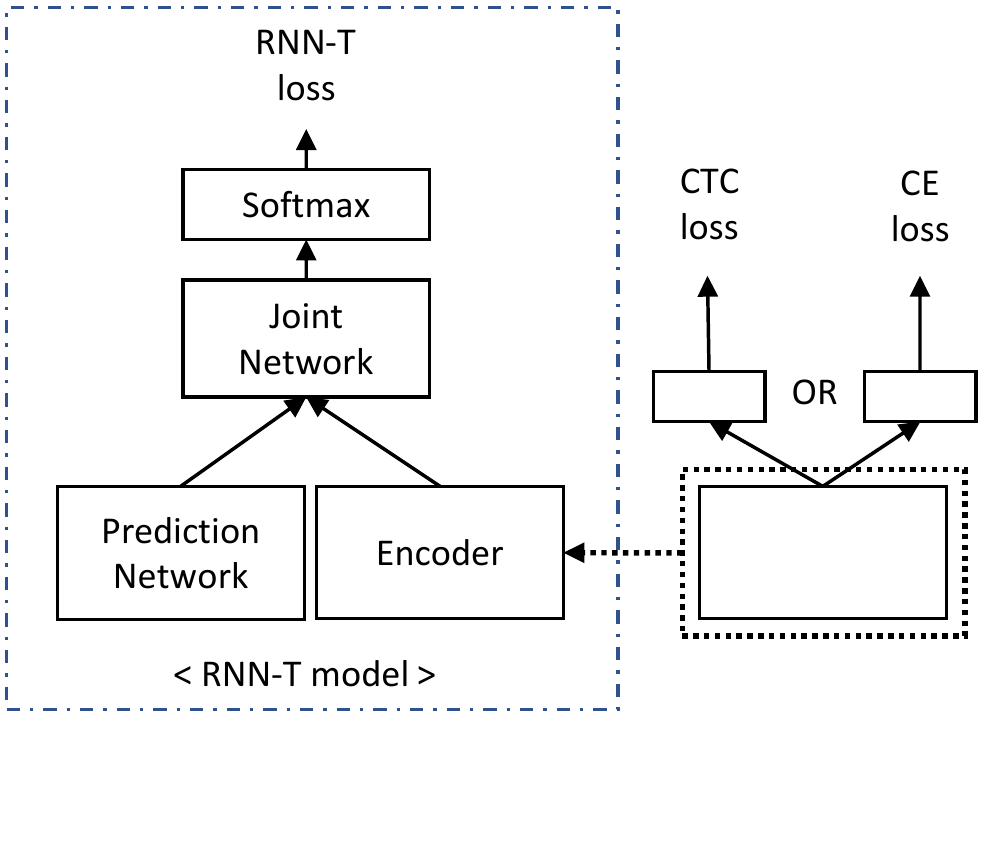}
  \vspace{-0.9cm}
  \caption{Illustration of encoder pre-training for RNN-T. Dashed arrow means initializing from a  pre-trained model.}
  \label{fig:init-sep}
\end{figure}

In our work, we propose to utilize external alignments to pre-train the encoder with the Cross Entropy (CE) criterion. The encoder is regarded as a token classification model rather than a CTC model. As shown in the right part of Figure~\ref{fig:init-sep}, an RNN based token classification model is trained first with the CE loss. In this paper we use 'CE loss' to represent the cross entropy loss function, and 'CTC loss' to represent the CTC forward-backward algorithm based loss function, and 'RNN-T loss' to represent the RNN-T loss function. 

In our experiments, we use word piece units as target tokens \cite{wu2016google}, which have been explored in the context of machine translation, and successfully applied in E2E ASR \cite{rnnt-google, ctc-ms}. With word-level alignments, we can get the boundary frame index of each word. For the word divided into more than one word piece, we equally allocate the total frames inside the word boundary to its word pieces. There will be a marginal case in which a word contains more word pieces than frames, which prevents us from generating token alignments. The total ratio of this special case is less than 0.01\% of all the training utterances, so we just remove those utterance in the pre-training stage. In this way, we can obtain the hard alignments of target tokens for all the frames.

Based on the encoder structure, one extra fully connected layer is added on top of the encoder, in which the output $h^{enc}_t$ is used for token classification. The objective is 
\begin{align}
L_{enc} = - \sum_{k=1}^K y_{t,k} * log\ (softmax\ (f^{fc}(h^{enc}_{t,k}))),
\end{align}
where $f^{fc}$ represents a fully connected layer, $k$ is the label index and $K$ denotes the target dimension, which is also the dimension of $z_{t,u}$. And $y_{t}$ is the word piece label for each input frame $x_t$. After the encoder pre-training, each output $h^{enc}_t$, which is the high-level representation of input acoustic features, is expected to contain the information about the alignments.

\section{whole-network pre-training}
\label{sec:init-com}
Among encoder pre-training methods, encoder is regarded to perform token mapping (CTC loss pre-training) or token aligning (CE loss pre-training). However, these pre-training methods only consider part of the RNN-T model. In this paper, we also explore the whole-network pre-training method with the use of external token alignments information. Different from other models, the output of the RNN-T is three-dimensional. Thus, the key challenge for the the whole-network pre-training is the label tensor $\mathbf{y}$ design. The optimizer reduces the CE loss between the output $z$ of the model and a crafted three-dimensional label tensor $\mathbf{y}$.

\begin{figure}[ht]
  \centering
  \includegraphics[width=1.0\linewidth]{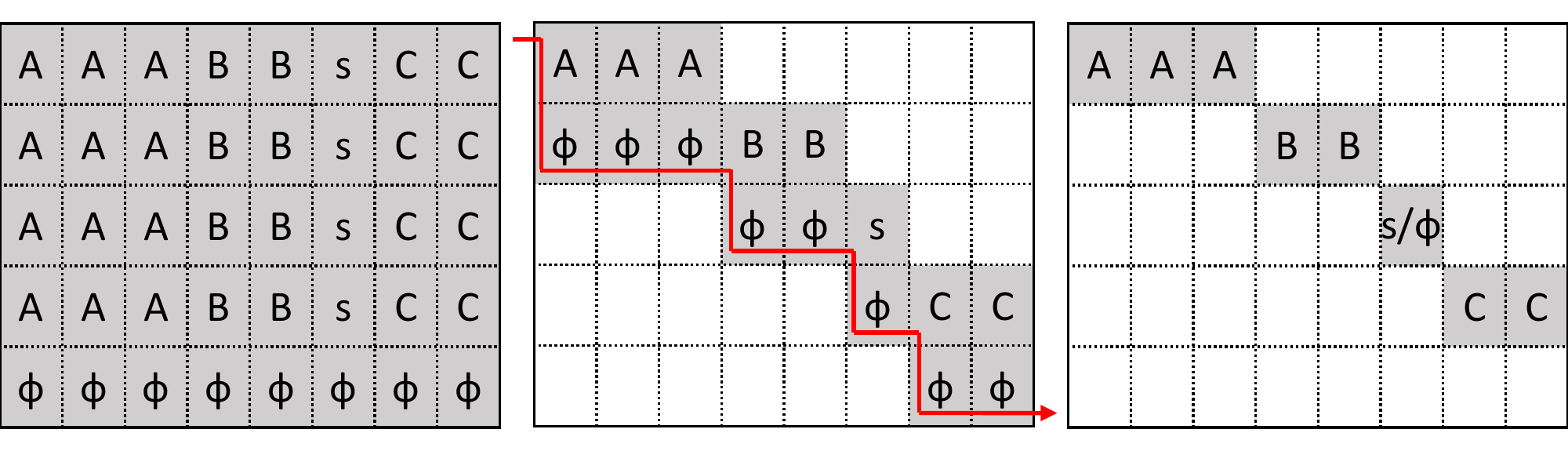}
  \vspace{-0.6cm}
  \caption{Examples of three designed label tensors for whole-network pre-training. Each grid represents an one-hot vector. The example 8-frame utterance is 'A B s C', and its alignment is 'A A A B B s C C'. In each label tensor, 's' means $space$ and '$\Phi$' means $blank$. From left to right, each sub-figure represents $\bf{y_1}$ to $\bf{y_3}$ as the order. Only gray grids are used for CE computing. The read arrow in $\mathbf{y_2}$ represents the decoding path when decoding.}
  \label{fig:init-com}
\end{figure}

Three different types of label tensor $\mathbf{y}$ are explored in this study, and examples are given in Figure~\ref{fig:init-com}. In each label tensor, the horizontal axis represents the time dimension, and the vertical axis represents the output token dimension. We represent $blank$ as an additional class, and it is shown as one-hot vector in the label tensor. At the first label tensor $\mathbf{y_1}$, we set all the output target grids of each frame in $\mathbf{y_1}$ to the one-hot vector corresponding to its alignment label. The last row of the label tensor is set to all $blank$, which indicates the end of the utterance. Thus, after pre-training, encoder output $h^{enc}_t$ is supposed to contain the alignment information. However, $\mathbf{y_1}$ only considers the frame-by-frame alignment, but ignores the output token level information. If we directly perform the RNN-T decoding on $\bf{y_1}$, we can not obtain the correct inference sequence.

Thus, taken the decoding process into consideration, we design another label tensor $\mathbf{y_2}$. Each frame is assigned to its token alignment. Target token position is determined by its sequence order. When perform pre-training, we only compute the CE from the non-empty part of the label tensor. $blank$ token is inserted under each target token to ensure the correct decoding results. If we directly perform the RNN-T decoding algorithm on the label tensor $\bf{y_2}$, correct results should be obtained. The decoding path is illustrated by the red arrow on the label tensor $\mathbf{y_2}$ in Figure~\ref{fig:init-com}. Thus, by directly performing the decoding on $\mathbf{y_2}$ of the given example, the inference result is 'A $\Phi$ $\Phi$ $\Phi$ B $\Phi$ $\Phi$ s $\Phi$ C $\Phi$ $\Phi$'. After removing $blank$ tokens, the final result is 'A B s C", which is the same with the alignment of this utterance.

However, in $\bf{y_2}$, almost half of the valid part is $blank$, so that $blank$ tokens dominate the pre-training process. Therefore, we design the label tensor $\mathbf{y_3}$, which only keeps the non-blank part of $\bf{y_2}$. The label tensor only remains one grid with its corresponding alignment for each frame. In order to provide the blank information during the pre-training stage, we set short-pause (space token less than 3 frames) of each utterance to $blank$. That means some $space$ in the valid part of the label tensor will become $blank$. After the pre-training is done, we replace the CE loss with the RNN-T loss, and proceed to the standard RNN-T training. 

\section{Experiments and Analysis}
\label{sec:exp}

\subsection{Experimental setup}
The proposed methods are evaluated on the 3400-hour Cortana voice assistant data, and 65,000-hour Microsoft production data.  For the Cortana data, the training and test sets consist of approximately 3400 hours and 6 hours of English audio, respectively. The 65,000 hours production data are transcribed data from all kinds of Microsoft products. The test sets cover 13 application scenarios such as Cortana and far-field speaker, with totally 1.8 million (M) words. Training and test material is anonymized, with personally identifiable information removed. In this work, we evaluate the methods on Cortana data at first, and then evaluate the selected best method on very large scale 65,000-hour production data.

The input feature is a vector of 80-dimension log Mel filter bank for every 10 milliseconds (ms) of speech. Eight vectores are stacked together to form an input frame to the encoder, and the frame shift is 30 ms. All RNN-T models adopt the configuration recommended in \cite{rnnt-ms, rnnt-device}. All encoders (Enc.) have 6 hidden-layer LSTMs, and all prediction networks (Pred.) have 2 hidden-layer LSTMs. The joint network has two linear layers without any activation functions. Layer normalization is used in all LSTM layers, and the hidden dimension is 1280 with the projection dimension equal to 640. The output layer models 4000 word piece units together with $blank$ token. The word piece units are generated by running byte pair encoding \cite{bpe} on the acoustic training texts.

\subsection{Evaluation on Cortana data}
Experimental results of whole-network pre-training are shown in the Table~\ref{tab:init-com}. The RNN-T baseline is trained from random initialization. For pre-trained models, the whole network is pre-trained with CE loss, then it is trained with RNN-T loss. Using the pre-trained network as a seed model, the final word error rate (WER) can be significantly reduced. All designed label tensors can improve the RNN-T training,  achieving 10\% to 12\% relative WER reduction.

\begin{table}[ht]
\caption{WER comparison of different whole-network pre-training methods on 3400 hours Cortana data. Pre-train (all align) uses $\mathbf{y_1}$, Pre-train (correct decoding) uses $\mathbf{y_2}$, and Pre-train (align path - sp blank) uses $\mathbf{y_3}$ as the target label sensors, respectively.}
\label{tab:init-com}
\vspace{0.2cm}
\centering
\begin{tabular}{l|c}
\hline\hline
Model                                   & WER(\%) \\ \hline\hline
RNN-T                          & 15.11   \\ \hline
+ Pre-train (all align)                 & 13.53   \\
+ Pre-train (correct decoding)        & 13.66   \\
+ Pre-train (align path - sp blank) & 13.23   \\ \hline\hline
\end{tabular}
\end{table}

In Table~\ref{tab:init-sep1}, we evaluate the encoder pre-training method on 3400 hours Cortana data. Using a pre-trained CTC model to initialize the encoder does not improve the accuracy. This is because the output of CTC is a sequence of spikes, in which there are lots of $blank$ tokens without any meaning. Hence, if we use the pre-trained CTC as the seed for the encoder of RNN-T, most encoder output $h^{enc}_t$ will generate $blank$, which does not help the RNN-T training. When we use CE loss pre-trained encoder to initialize the encoder of RNN-T, it achieves significant improvement compared with training from the random initialization. It obtains 28\% relative WER reduction from the RNN-T baseline and CTC based encoder pre-training.

\begin{table}[ht]
\caption{WER comparison of different encoder pre-training methods on 3400 hours Cortana data. Greedy search is used. 'CTC' means pre-training of CTC loss with target sequence, 'CE' means pre-training of CE loss with target alignment. 'no' means training from the random initialization.}
\label{tab:init-sep1}
\vspace{0.2cm}
\centering
\begin{tabular}{l|c|c}
\hline\hline
Model          & Enc. Pre-train & WER (\%)\\
\hline\hline
RNN-T baseline & no                                                       & 15.11    \\\hline
+ Pre-train    & CTC                                                      & 15.07                                              \\
+ Pre-train    & CE                                                       & 10.83   \\ \hline\hline                                          
\end{tabular}
\end{table}

Among all the encoder pre-training experiments in Table~\ref{tab:init-sep1}, prediction network and joint network are all trained from the random initialization. The only difference is the parameters seed of encoder. When comparing CTC loss and CE loss based encoder pre-training methods, there is a huge WER gap between these two approaches. Initializing the encoder as a token aligning model rather than a sequence mapping model results in the much better accuracy. This is because the RNN-T encoder performs the frame-to-token aligning, which extracts the high-level features of each input frame.

\subsection{Evaluation on very large scale data}
From our experiments, both the encoder pre-training and the whole-network pre-training can improve the performance of RNN-T model. In order to get more convincing results, we evaluate our proposed methods on very large scale data, where we use the 65,000 hours Microsoft production data set. The results are shown in Table~\ref{tab:init-sep2}. Due to the very large resource requirement and computation cost, we only evaluate CE-based encoder pre-training method, which obtained the best accuracy in Cortana experiments. All the results are obtained using beam search, and the beam width is 5.

Besides our proposed methods, we evaluate the widely used CTC-RNNLM pre-training strategy \cite{rnnt-google, rnnt-xielei, graves2013speech} as comparison. It used a well trained CTC model to initialize the encoder, and a well trained RNNLM to initialize the prediction network. This CTC+RNNLM initialization approach reduced the average WER from 12.63 to 12.29 in 13 test scenarios with 1.8 M words. In contrast, our proposed approach, which pre-trains the encoder with alignments using the CE loss, outperforms the other methods significantly, achieving a 11.41 WER on the average. Compared with training from the random initialization, our proposed method can obtain 10\% relative WER reduction in the very large scale task.

\begin{table}[ht]
\caption{WER comparison of different encoder pre-training methods on the task with 65,000 hours production training data.}
\label{tab:init-sep2}
\vspace{0.2cm}
\centering
\begin{tabular}{l|c}
\hline\hline 
Model          &  WER (\%) \\
\hline\hline 
RNN-T baseline                                                                                                          & 12.63                                              \\ \hline
+ Pre-train (Enc. CTC, Pred. LM) \cite{rnnt-google}                                                                                                        & 12.29                                              \\
+ Pre-train (Enc. CE)                                                                                                       & 11.34    \\ \hline\hline                                          
\end{tabular}
\end{table}

\subsection{Output time delay comparison}
\begin{figure}[ht]
  \centering
  \includegraphics[width=1.0\linewidth]{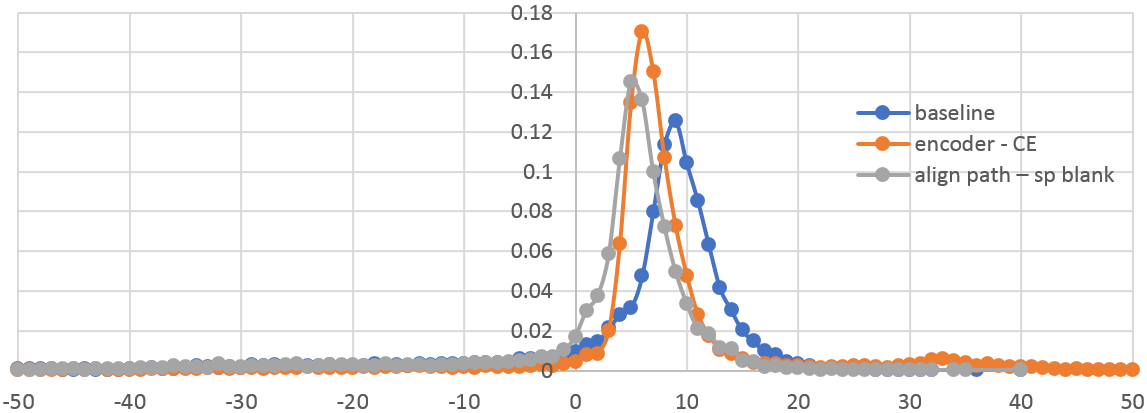}
  \vspace{-0.2cm}
  \caption{Frame delay difference between ground truth word alignment and the word alignment generated from different RNN-T models.}
  \label{fig:delay}
\end{figure}

Although RNN-T is a natural streaming model, it still has latency compared to hybrid models \cite{rnnt-ms}. With the help of alignments for model initialization, we hope to reduce the latency of RNN-T. To better understand the advantages of our proposed pre-training methods, we compare the gap between the ground truth word alignment and the word alignment generated by greedy decoding from different RNN-T models. The visualization is performed on the test set of Cortana data. As shown in the Figure~\ref{fig:delay}, the central axis represents the ground truth word alignment. The output alignment distributions are normalized to the normal distribution. The horizontal axis represents the number of frames away from the ground truth word alignment, and the vertical axis represents the ratio of words. 

\vspace{0.0cm}

From the Figure~\ref{fig:delay}, different RNN-T models have different time delay compared with the ground truth. That's because the RNN-T model tends to see several future frames, which can provide more information for the token recognition. The baseline RNN-T model has around 10 frames average delay. In contrast, when performed the proposed pre-training methods, the average delay can be significantly reduced. Using CE pre-trained encoder to initialize the RNN-T model can reduce the average delay to 6 frames, and using whole-network pre-training method can reduce it to 5 frames. The reason for the time delay reduction is that pre-training provides the alignment information to the RNN-T model, which will guide the model to make decision earlier. This shows the advantage of our proposed pre-training methods in terms of time delay during the decoding stage. 

\vspace{-0.3cm}
\section{Conclusion}
\label{sec:con}
In this work, we explore the training strategy of an RNN-T model and propose two pre-training approaches with the use of external alignment information. Two types of pre-training methods have been evaluated, referred to as encoder pre-training and whole-network pre-training. Encoder pre-training used CE loss to pre-train the encoder of RNN-T only. Whole-network pre-training pre-trains the whole RNN-T model with CE loss. Three kinds of designed label tensors are used for the whole-network pre-training. The proposed methods are evaluated on 3400 hours Cortana data and 65000 hours production data. When compared with training from the random initialization, the whole-network pre-training obtains a 12\% relative WER reduction. And the encoder pre-training obtains a 28\% and a 10\% relative WER reduction on 3400 hours Cortana and 65,000 hours production data, respectively. Compared to the widely used CTC+RNNLM initialization strategy on very large scale data, encoder pre-training still outperforms it by a 8\% relative WER reduction. Our proposed methods can also significantly reduce the time delay of RNN-T model.

\clearpage
\begin{spacing}{0.88}
\bibliographystyle{IEEEbib}
\bibliography{refs}
\end{spacing}

\end{document}